\documentclass[11pt,a4paper]{article}
\usepackage[hyperref]{eacl2021}
\usepackage{times,latexsym}

\usepackage{microtype}

\usepackage{graphicx}
\graphicspath{ {./imgs/} }
\usepackage{caption}
\usepackage{subcaption}
\usepackage{amsmath}
\usepackage{amssymb}
\usepackage{amsfonts}
\usepackage{bbm}
\usepackage{rotating}
\usepackage{booktabs}
\usepackage{multicol}

\usepackage{cleveref}
\crefname{section}{\S}{\S\S}
\Crefname{section}{\S}{\S\S}
\crefname{table}{Tab.}{}
\crefname{figure}{Fig.}{Figs.}
\crefname{algorithm}{Algorithm}{}
\crefname{algorithm}{Algorithm}{}
\crefname{line}{Line}{}
\crefname{appendix}{App.}{}
\crefformat{section}{\S#2#1#3}
\crefname{thm}{Theorem}{}
\crefname{prop}{Proposition}{}
\crefname{def}{Definition}{}

\DeclareMathOperator{\ent}{H}
\DeclareMathOperator{\mi}{MI}

\newcommand{\enttheta}{\ent_{\theta}}

\DeclareMathOperator{\countop}{count}
\DeclareMathOperator{\softmax}{softmax}

\DeclareMathOperator{\LSTM}{LSTM}
\DeclareMathOperator{\KL}{KL}
\DeclareMathOperator*{\Expect}{\mathbb{E}}

\newcommand{\mask}{\mathbf{MASK}}
\newcommand{\trie}{q_{\mathrm{trie}}}
\newcommand{\eow}{\textsc{eow}}

\newcommand{\bw}{\mathbf{w}}
\newcommand{\bz}{\mathbf{z}}
\newcommand{\bh}{\mathbf{h}}

\newcommand{\word}[1]{\textit{#1}}
\newcommand{\defn}[1]{\textbf{#1}}

\newcommand\citepossessive[1]{\citeauthor{#1}'s\ (\citeyear{#1})}

\aclfinalcopy %
\def\aclpaperid{16} %

\everypar{\looseness=-1}

\title{Disambiguatory Signals are Stronger in Word-initial Positions}
\usepackage{tipa}

\newcommand{\ucambridge}{\normalfont \text{\textipa{D}}}
\newcommand{\ethz}{\text{\normalfont \textipa{Q}}}
\newcommand{\google}{\normalfont \text{\textipa{@}}}

\author{Tiago Pimentel$^{\ucambridge}$ \And
Ryan Cotterell$^{\ucambridge,\ethz}$ 
\\
  $^{\ucambridge}$University of Cambridge~\;~\;~%
  $^{\ethz}$ETH Z\"{u}rich~\;~\;~%
  $^{\google}$Google\\
  \texttt{tp472@cam.ac.uk},~\;~ \texttt{ryan.cotterell@inf.ethz.ch},~\;~ \texttt{roark@google.com}
  \And
Brian Roark$^{\google}$
}

\date{}

\begin{document}
\maketitle
\begin{abstract}
Psycholinguistic studies of human word processing and lexical access provide ample evidence of the preferred nature of word-initial versus word-final segments, e.g., in terms of attention paid by listeners (greater) or the likelihood of reduction by speakers (lower).
This has led to the conjecture---as in \newcite{wedel2019incremental}, but common elsewhere---that languages have evolved to provide more information earlier in words than later.  Information-theoretic methods to establish such tendencies in lexicons have suffered from several methodological shortcomings that leave open the question of whether this high word-initial informativeness is actually a property of the lexicon or simply an artefact of the incremental nature of recognition.  In this paper, we point out the confounds in existing methods for comparing the informativeness of segments early in the word versus later in the word, and present several new measures that avoid these confounds.  When controlling for these confounds, we still find evidence across hundreds of languages that indeed there is a cross-linguistic tendency to front-load information in words.\footnote{Our code is available at \url{https://github.com/tpimentelms/frontload-disambiguation}.}
\end{abstract}

\section{Introduction}

The psycholinguistic study of human lexical access is largely concerned with the incremental processing of words---whereby, as individual sub-lexical units (e.g., phones) are perceived, listeners update their expectations of the word being spoken. One common tenet of such studies is that the disambiguatory signal contributed by units early in the word is stronger than that contributed later---i.e. \defn{disambiguatory signals are front-loaded in words}.
This intuition is derived from ample indirect evidence that the beginnings of words are more important for humans during word processing%
---including, e.g.,
evidence of increased attention to word beginnings \cite[\textit{inter alia}]{nooteboom1981lexical} or evidence of increased levels of phonological reduction in word endings \cite{son2003efficient}.

\begin{figure}
    \centering
    \includegraphics[width=\columnwidth]{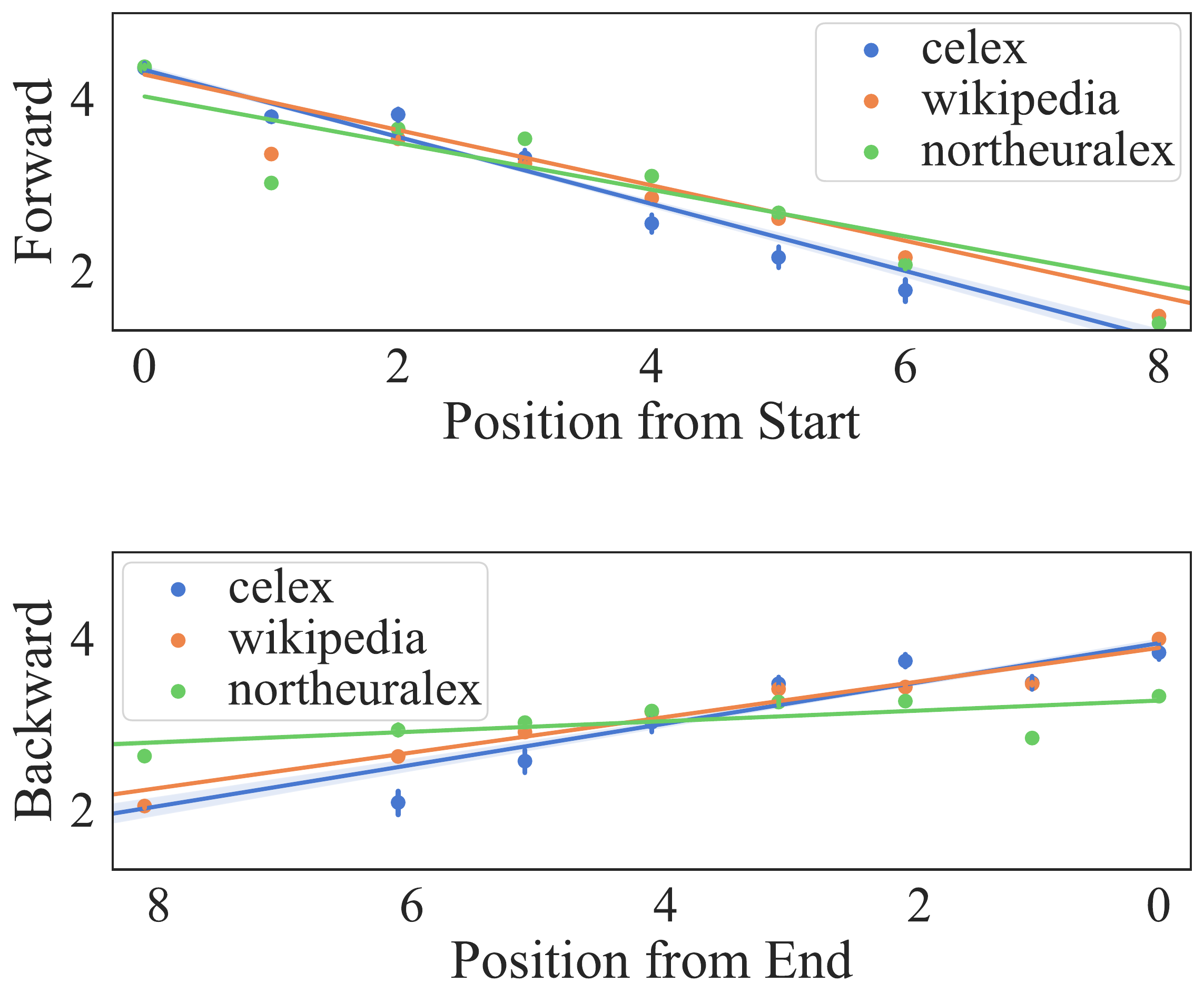}
    \caption{\small Forward and Backward Surprisals with LSTM model from \citet{pimentel2020phonotactic}. The bottom plot has been flipped horizontally such that it visually corresponds to the normal string direction.}
    \label{fig:surprisals_forward_backward--lstm}
\end{figure}

To analyse this front-loading effect, researchers have investigated the information provided by segments in words.
\newcite{son2003information,son2003efficient}  showed that, in Dutch, a segment's position in a word is a very strong predictor of its conditional surprisal,
with later segments being more predictable than earlier ones---a result which we show to arise directly from its definition in \cref{sec:trivial}.
Recently \newcite{king2020greater} and \newcite{pimentel2020phonotactic} confirmed the effect on many more languages.

Their analysis, however, presents an inherent confound between the amount of conditional information available to a model and the surprisal of the subsequent segment---see \cref{fig:surprisals_forward_backward--lstm} for results illustrating this.
Using the LSTM training recipes from \citet{pimentel2020phonotactic},\footnote{https://github.com/tpimentelms/phonotactic-complexity} we calculated the conditional surprisal at each segment position within the words across all languages in three datasets.\footnote{See \cref{sec:methods} and \cref{sec:data} for specifics on training and data. Each segment corresponds to a single phone in CELEX and NorthEuraLex, and to a single grapheme in Wikipedia.} The top-half of \cref{fig:surprisals_forward_backward--lstm} shows that, indeed, positions earlier in the string have higher surprisal than positions later in the string, supporting the thesis of higher informativity earlier in words. The bottom-half shows that modelling the strings right-to-left instead of left-to-right reverses the resulting effect.
This decouples conditional surprisal from the disambiguatory strength. %
To expose this decoupling, consider an artificial language where every word contains a copy of its first half, e.g., \word{foofoo}, \word{barbar}, \word{foobarfoobar}, etc.
The first and second halves of these words have identical disambiguatory strength; they are the same so one could disambiguate the word as easily from its second half as from the first. 
In contrast, conditional surprisal would be nearly zero for the second halves of words because the second half is perfectly predictable from the first half.\looseness=-1

In natural languages, measuring conditional entropy in a left-to-right fashion inherently forces a reduction of conditional entropy in later segments because of a language's phonotactic constraints.
However, the disambiguatory strength of later segments is not inherently less than that of earlier segments.
For instance, in a language like Turkish, which has vowel harmony, knowledge of any of the vowels in a word will provide information about the word's other vowels in a similar way. 
As such, knowledge of vowels towards the front of a word is as disambiguating as of vowels towards its end.

The contributions of this paper are threefold.  First, we document and demonstrate the shortcomings of existing methods for measuring the informativeness of individual segments in context, including the confound with the amount of conditional information discussed above. Second, we introduce three surprisal-based measures that control for this confound and enable comparison of word-initial versus -final positions in this respect: unigram, position-specific and cloze surprisal (see \cref{sec:methods}).
Finally, we find robust evidence across many languages of 
stronger disambiguatory signals in word initial than word-final positions. Out of a total of 151 languages analysed across three separate collections, 82 of them present a higher cloze surprisal in word beginnings than in endings—with similar patterns arising with the other two measures.
\section{Background and Related Work}\label{sec:critique}

\paragraph{Psycholinguistic evidence.} Lexical access has long been a topic of interest for psycholinguists, leading to many distinct models being proposed for this process \citep{morton1969interaction,marcus1981eris,marslen1987functional}. Far earlier, though, \citet{bagley1900apperception} had already demonstrated that earlier segments in words were more important for word recognition than later segments; specifically, they found that, when exposed to words with word-initial or word-final consonant deletions, listeners found the word-initial deletions more disruptive.
\citet{fay1977malapropisms} showed mispronunciations are more likely in word endings, while \citet{bruner1958note} showed that recognizing written words with flipped initial characters was harder than with word final ones---demonstrating that the initial part of the word was more ``useful'' for readers. More recently, \citet{wedel2019crosslinguistic} found evidence in support of \citet{houlihan1975role}, showing neutralizing rules tend to target word endings more significantly than beginnings in both suffixing and prefixing languages.

\citet{nooteboom1981lexical} investigated the ease of recovering lexical items from either word beginnings or endings, finding that people had an easier time recovering words from their beginnings. For this, he examined words for which the first and second halves each completely identified them in a large Dutch dictionary---controlling for both segments' length and uniqueness.
Later on, though, \citet{nooteboom1988search} showed this difference vanishes when priming people with the length of the word---proposing the difference comes not from how informative segments were, but from the difficulty in time aligning later segments in mental lexicons. \citet{connine1993beginnings} also found no difference in priming effects with non-words that differed from real words in either word initial or medial positions, suggesting initial positions have no special status in word recognition.

Psycholinguistic evidence is key to understanding how lexical access works in human language processing, and can help us understand why lexicons may evolve to provide more disambiguatory signals earlier in words.\footnote{Note that there are many possible reasons why the effects we demonstrate in this paper may arise, from the demands of lexical access to constraints on articulation. We provide no evidence for any of the possible explanations, evolutionary or otherwise, just methods for measuring the effect.} Given the incremental nature of human lexical processing, however, such evidence cannot provide direct evidence of the nature of the lexicon uninfluenced by incrementality.

\paragraph{Computational evidence.} To the best of our knowledge, \citet{son2003efficient,son2003information} were the first to use computational methods coupled with an information theoretic definition of informativeness to investigate this question. They showed that segments in the beginning of words carry most of a word's information, as measured by their contextual surprisal using a plug-in tree structured probabilistic estimator.  Although assessing a less-biased sample of words than \citet{nooteboom1981lexical},\footnote{\citet{nooteboom1981lexical} looked at words completely identifiable by both their first and second halves in a large Dutch dictionary---this resulted in a study with only 14 words.}
 this study is also limited to a single language (Dutch), hence cannot assess whether this is a general phenomenon or specific to that language.

Further, \citet{son2003information,son2003efficient} use absolute word positions in their analysis. Word length correlates strongly with frequency, hence while early positions are present in all words, later positions only exist for a much smaller sample of typically lower frequency words. Thus this comparison amounts to asking if later positions in longer and infrequent words have lower surprisal than earlier positions in all (frequent or infrequent) words. We analyse this confounding factor in \cref{sec:study_length}.

\citet{wedel2019incremental} and \citet{king2020greater} applied a methodology similar to that of \citet{son2003information} to show, for many diverse languages, that more frequent words contain less informative segments in word initial positions, while less frequent types
carry more informative ones. 
They further showed that segments in later word positions were less informative (given the previous ones) than average in rarer words. 
While controlling for length, \citet{king2020greater} also compared words' forward and backward uniqueness points---nodes in a trie from which only one leaf node can be reached, i.e., where the word is uniquely identified---showing they happened earlier in forward strings.

While these studies provide evidence from more diverse sets of languages, they follow \citet{son2003information} in studying closed lexicons.\footnote{The closed lexicon assumption is incorporated implicitly in the probabilistic trie models used by \citet{son2003information,son2003efficient} and \citet{king2020greater}---i.e. they assign zero probability to any form not in their training sets---and in the uniqueness point analysis of \citet{king2020greater}.} As we show in \cref{sec:trivial}, 
the use of probabilistic trie models on a closed lexicon yields a trivial effect of higher informativity at word initial positions.
Furthermore, such studies cannot account for out-of-vocabulary words (e.g., nonce, proper name or otherwise unknown words) or derivational morphology, which are key parts of lexical recognition.
Lexical access is also somewhat robust to segmental misordering \cite{toscano2013reconsidering} and sounds later in a word help determine the perception of earlier ones \cite{gwilliams2018spoken}.  In contrast, a trie over a closed lexicon is deterministic.
Beyond this, 
\citet{luce1986computational} showed in a corpus study that the probability of a word type being uniquely identifiable before its last segment was only $41\%$---and $19\%$ of types were identified only by the end of word, being proper prefixes of other words, such as \emph{cat} and \emph{cats}. They conclude that uniqueness point statistics may only be useful for long word analysis.

In \citet{pimentel2020phonotactic}, we analysed several languages' phonotactic distributions, focusing on presenting a trade-off between phonotactic entropy and word length across languages. As a control experiment we analysed the correlation between a segment's surprisal and its word position across 106 languages. We did not control for word length and did not run per-language experiments, though---so we could have just been capturing the effect that later positions will mostly be present in languages with longer words (which, as we find, have lower information on average).\footnote{We note this issue only applies to the control experiment, and has no bearing on the key findings of that paper.}

While this last work avoids many of the issues raised earlier in this section, it fails to control the key confound mentioned earlier: it relies on left-to-right conditional probabilities to calculate surprisal. Thus segments early in the word have less conditional information and hence are generally of lower probability---a trivial effect that does not indicate a segment's disambiguatory signal strength.

\section{Measures of Disambiguatory Strength}\label{sec:methods}
\subsection{A Lexicon Generating Distribution}
In this work, instead of the lexicon itself, we investigate the probability distribution from which it is sampled. 
The distribution is unobserved, but we can get glimpses of it via the sampled lexicon:
\begin{equation}
    \left\{\bw^{(n)}\right\}_{n=1}^N \sim p(\bw) = \prod_{t=1}^{|\bw|} p(w_t \mid \bw_{< t})
\end{equation}
\noindent 
The distribution $p(\bw)$ is defined over the entire space of possible phonological wordforms
$\bw \in \Sigma^*$, where $\Sigma$ is a language-specific alphabet
and the operator $^*$ indicates its Kleene closure.\footnote{
We pad all strings with the end-of-word (\eow{}) symbol. For simplicity, we assume the alphabet includes \eow{} throughout the rest of the paper.
}
This distribution should assign high probability to likely wordforms (attested or not) and low probability to unlikely ones. Using \citepossessive{chomsky1965some} classic example from English, \word{brick} (attested) and \word{blick} (unattested) would have high probability, whereas \word{*bnick} (unattested) would have a low probability.

\subsection{Entropy and Conditional Entropy}
Shannon's entropy is a measure of how much information a random variable contains.
Consider a segment $w_t$ at word position $t$, which is a value of the random variable $W_t$.
The average information (surprisal) relayed per segment is:
\begin{equation}
    \ent(W_t) \equiv \sum_{w_t \in \Sigma} p(w_t) \log \frac{1}{p(w_t)}
\end{equation}
A random variable is maximally entropic if it is a uniform distribution, in
which case $\ent(W_t) = \log(|\Sigma|)$. 
Conditional entropy measures how much information the knowledge of a variable conveys, given some previous knowledge. The average information transmitted per segment, given the previous ones in a word, is
\begin{align}
    \ent(W_t \mid &W_{<t}) \equiv \\
    &\sum\limits_{\bw_{\leq t} \in \Sigma^*} p(\bw_{\leq t}) \log \frac{1}{p(w_t \mid \bw_{<t})} \nonumber
\end{align}
where $\bw_{\leq t} = \bw_{< t} \circ w_t$.
We note the conditional entropy is always smaller or equal to the entropy, i.e. $\ent(W_t \mid W_{<t}) \leq \ent(W_t)$.\looseness=-1

\subsection{Plug-in Estimators, Context Size, and Disambiguatory Strength}

Our criticism of previous work investigating the disambiguatory strength of word-initial vs. word-final segments can be mainly divided in two parts: (i) the use of maximum likelihood plug-in estimators of the conditional entropy, by e.g. \citet{son2003efficient}; (ii) the use of left-to-right conditional entropy in itself, by all previous information-theoretic work in this vein.

\subsubsection{A Critique of \newcite{son2003efficient}}\label{sec:trivial}
We present a \textit{reductio ad absurdum} which shows that
\citepossessive{son2003efficient} method will lead to the conclusion that word-initial segments are more informative even if all segments were equally entropic and sampled independently---a nonsensical finding.
Accordingly, assume the probability distribution $p(w_t \mid \bw_{<t})$, from which each segment in a word is sampled, was independent, e.g. define
\begin{equation}
    \hat{p}(\bw) = \prod_{t=1}^{|\bw|} \hat{p}(w_t \mid \bw_{< t}) = \prod_{t=1}^{|\bw|} \hat{p}(w_t)
\end{equation}
Assume now that a large, but finite, lexicon is sampled from it $\left\{\hat{\bw}^{(n)}\right\}_{n=1}^N \sim \hat{p}(\bw)$. 
Further consider modelling this sampled lexicon with a probabilistic trie structure, similarly to what was done by \citet{son2003information,son2003efficient},\footnote{This is in fact a simplification of \citepossessive{son2003information} model, which in practice uses Katz smoothing.} i.e.
\begin{equation}
    \trie(w_t \mid \bw_{< t}) = \frac{\countop(w_t, w_{t-1}, \ldots, w_0)}{\countop(w_{t-1}, \ldots, w_0)}
\end{equation}
where $w_0$ is the beginning-of-word symbol.
Such a model uses all $N$ words to approximate the distribution of the first segment---i.e. $\countop(w_0) = N$.
Yet after $t-1$ segments, an exponentially smaller sample is used
to capture the distribution---i.e. $\Expect[\countop(w_{t-1}, \ldots, w_0)] = N / |\Sigma|^{t-1}$.
Using this model as a plug-in estimator of the entropy will lead to negatively biased estimates, where the error is approximately \citep{basharin1959statistical}:
\begin{align}
    \mathrm{H}(W_t \mid W_{t-1}&) - \Expect \left[ \hat{\mathrm{H}} \right]  \approx \frac{ (|\Sigma| - 1)\, \log e}{\countop(w_{t-1}, \ldots, w_0)} \nonumber \\
    &\approx \frac{ |\Sigma|^{t-1} (|\Sigma| - 1)\, \log e}{N}
\end{align}
where $\hat{\mathrm{H}}$ is a plug-in estimate of the entropy.
The error grows exponentially in $t$ due to the $|\Sigma|^{t-1}$ 
factor.
However, by assumption, $\ent(W_t \mid W_{t-1})$ is constant---we have equally entropic and independent segments.
Thus, the only way for this difference to increase is for the second term to decrease as a function of $t$. 
It follows that the estimated cross-entropies decrease as a function of $t$ due to a methodological technicality.
Indeed, in the extreme case, every position after a word's uniqueness point
would be estimated to have zero entropy.
Thus, \citepossessive{son2003information} method only reveals a trivial effect.\looseness=-1

\subsubsection{Conditional Entropy and Context Size}

As previously mentioned, the conditional entropy measures how much information the knowledge of a variable conveys, given some previous information, and it is always smaller or equal to the entropy.
For this reason, relying on left-to-right conditional entropies to estimate the strength of disambiguatory signals yields straightforward results; the availability of larger conditioning contexts in a word's final segments will naturally reduce its conditional entropy. This will negatively skew the estimated informativeness of the later parts of a word.
\begin{align}
    \ent(W_t) &\geq \ent(W_t \mid W_{t-1}) \geq \ent(W_t \mid W_{<t})
\end{align}

This effect can also be easily demonstrated by the symmetrical nature of mutual information (MI), where the MI is defined as:
\begin{align}
    \mi(W_t; W_{t-1}) &= \ent(W_t) - \ent(W_t \mid W_{t-1}) \nonumber\\ 
    & = \ent(W_{t-1}) -  \ent(W_{t-1} \mid W_t) \nonumber \\
    & = \mi(W_{t-1}; W_t) 
\end{align}
If we assume both segments had the same unconditional entropy, i.e. $\ent(W_t) = \ent(W_{t-1})$, then using left-to-right conditional entropies would suggest the later segment was less informative, while right-to-left conditioning would imply the opposite.
Nonetheless, both their contextual and uncontextual disambiguatory strength would in fact be the same, if we estimated it with equal-sized contexts:
\begin{align}
\ent(W_t) = \ent(&W_{t-1}) \Longrightarrow \\
    &\ent(W_t \mid W_{t-1}) = \ent(W_{t-1} \mid W_t) \nonumber
\end{align}

\subsection{Cross-Entropy and Entropy}

As mentioned above, the distribution $p(\bw)$ is not directly observable. We can, however, approximate it using character-level language models $p_\theta(\bw)$.
We are interested in the entropy of variable $W_t$, as a proxy we measure its cross-entropy
\begin{equation}
    \ent_{\theta}(W_t) \equiv \sum\limits_{w_t \in \Sigma} p(w_t) \underbrace{\log \frac{1}{p_{\theta}(w_t)}}_{\mathrm{surprisal}}
    \label{eq:cross_entropy_orig}
\end{equation}
where the surprisal is the information provided by a single segment instance $w_t$.
The cross-entropy is an upper bound on the entropy, i.e. $\ent(W_t) \le \ent_{\theta}(W_t)$, with their difference being the Kullback--Leibler (KL) divergence between both distributions.
Since the KL-divergence is always positive, this upper-bound holds. Furthermore, the closer $p_{\theta}$ is to the true distribution $p$, the smaller the divergence is, and the tighter this bound. As such, the better our model is at estimating the true distribution, the better our estimates of the entropy will be.

Calculating \cref{eq:cross_entropy_orig} still requires knowledge of the true $p$. We overcome this limitation by empirically estimating it on a held out part of the lexicon%
\begin{align}
    \ent_{\theta}(W_t) &\approx \frac{1}{N}\sum\limits_{n = 1}^N \log \frac{1}{p_{\theta}(w_t^{(n)})}
    \label{eq:cross_entropy}
\end{align}

\subsection{Earlier vs. Later Word Entropy}

For the remainder of this work, we will discuss information in terms of surprisal, since the entropy is its expected value.
We analyse the distribution of disambiguatory information across word positions via three distinct  measures---all of which control for the amount of conditioning per position:
\begin{itemize}
  \setlength\itemsep{0em}
    \item \defn{Unigram Surprisal $\enttheta(W_t)$:} the surprisal of individual segments.
    \item \defn{Cloze Surprisal $\enttheta(W_t \mid W_{\ne t})$:} surprisal of a segment given all others in the same word.
    \item \defn{Position-Specific Surprisal} ~\\ $\enttheta(W_t \mid T=t, |W|)$: the surprisal of individual segments given their position in the wordform and the word's length.
\end{itemize}

The unigram surprisal captures the information provided by each segment when considering no context; while the cloze surprisal represents the information provided by a segment when one already knows the rest of the word. The position-specific surprisal represents a mid way between both, conditioning each segment only on its position and the word's length---being inspired by \citepossessive{nooteboom1988search} experiments.
These three measures of information control for the context size considered at each position, being thus better for an investigation of disambiguatory strength.

We used an unigram model (see \cref{sec:models}) to estimate the unigram surprisal, and transformers \citep{vaswani2017attention} for cloze and position-specific surprisals.
We also use the LSTM \citep[Long-Short Term Memory,][]{hochreiter1997long} model from \newcite{pimentel2020phonotactic} for two other entropy measures which do not control for the amount of conditional information:

\begin{itemize}
  \setlength\itemsep{0em}
    \item \defn{Forward Surprisal $\enttheta(W_t \mid W_{<t})$:} the surprisal of a segment given the previous ones.
    \item \defn{Backward Surprisal $\enttheta(W_t \mid W_{>t})$:} the surprisal of a segment given the future ones.
\end{itemize}

We include the beginning- and end-of-word symbols in the forward and backward surprisal analysis, respectively, following previous work \citep{wedel2019incremental,pimentel2020phonotactic,king2020greater}. However, we ignore them in the unigram, position-specific and cloze surprisal analyses. Position-specific and cloze surprisal are given information about word length, hence these symbols are unambiguously predictable. We analyse the impact of these symbols in \cref{sec:eow}.

\section{Character-Level Language Models}\label{sec:models}
In this paper, we make use of character-level language models to model the probability distributions $p_{\theta}$ and approximate the relevant cross-entropies.

\paragraph{Unigram.}
This might be the simplest language model still in use in Natural Language Processing. We use its Laplace-smoothed variant
\begin{equation}
    p_{\theta}(w_t) = \frac{\countop(w_t) + 1}{\sum_{c' \in \Sigma} \countop(c') + |\Sigma|}
\end{equation}

\paragraph{LSTM.}
This architecture is the state-of-the-art for character-level language modelling \citep{melis2020mogrifier}.
Given a sequence of segments $\bw \in \Sigma^*$, we use one hot lookup embeddings to transform each of them into a vector $\bz_t \in \mathbb{R}^d$. We then feed these vectors into a $k$-layer
LSTM
\begin{equation}
    \bh_t = \LSTM(\bz_{t-1}, \bh_{t-1})
\end{equation}
where $\bh \in \mathbb{R}^d$, $\bh_0$ is a vector with all zeros and $w_0$ is the beginning-of-word symbol. We then linearly transform these vectors before feeding them into a softmax non-linearity to obtain the %
distribution
\begin{equation}
    p_{\theta}(w_t \mid \bw_{<t}) = \softmax(W \bh_t + b)
\end{equation}
in this equation, $W \in \mathbb{R}^{|\Sigma| \times d}$ is a weight matrix and $b \in \mathbb{R}^{|\Sigma|}$ a bias vector.

\paragraph{Backward LSTM.}
To get the backward surprisals we use models with the same architecture, but reverse all strings before feeding them to the models. As such, we get the similar equations
\begin{align}
    \bh_t &= \LSTM(\bz_{t+1}, \bh_{t+1}) \\
    p_{\theta}(w_t \mid \bw_{>t}) &= \softmax(W \bh_t + b)
\end{align}

\paragraph{Transformer.}
Transformers allow a segment to be conditioned on both future and previous symbols. Our implementation starts similar to the LSTM one, getting embedding vectors $\bz_t$ for each segment in the string $\bw \in \Sigma^*$, except that we replace segment $w_t$ with a $\mask$ symbol. We then feed these vectors through $k$ multi-headed self-attention layers, as defined by \citet{vaswani2017attention}.
Finally, the representations from the last layer are  linearly transformed and fed into a softmax
\begin{equation}
    p_{\theta}(w_t \mid \bw_{\ne t}) = \softmax(W \bh_t + b)
\end{equation}

\paragraph{Position-Specific Transformer.}
To get position-specific surprisal values, we again use a transformer architecture, but instead of replacing a single segment with a $\mask$ symbol, we replace all of them. This is equivalent to conditioning each segment's distribution on its position and the word length---i.e., estimating $p_{\theta}(w_t \mid t, \left|\bw\right|)$.
\section{Data}\label{sec:data}

In order to estimate redundancy and informativeness of segments we use three different datasets, each with its own pros and cons. 
We focus on types instead of tokens---i.e., the datasets consist of lexicons---for a few different reasons. First, it is easier to get reliable samples of types than tokens for a language, specially low-resource ones.
Second, it is a well known result that token frequency correlates with both word length \citep{zipf1949human} and phonotactic probability \citep{mahowald2018word,meylan2017word},
so that would be a strong confound in the results. Third, morphology is more easily modeled at the type level than at token level \citep{goldwater2011producing}.%
\footnote{For each of the analysed datasets, we use 80\% of the word types for training, with the rest being equally split between development and test sets; only test set surprisal and cross-entropies are used in our analysis.}

\textbf{CELEX} \citep{baayen2014celex2} allows us to experiment exclusively on monomorphemic words, but covers only three closely related languages.
It contains both morphological and phonetic annotations for a large number of words in English, Dutch and German. We follow \citet{dautriche2017words} in using only words labeled as monomorphemic in our study, leaving us with $4{,}810$ words in German, 
$6{,}206$ words in English and $7{,}045$ words in Dutch.

\textbf{NorthEuraLex} \citep{dellert2019northeuralex} spans 107 languages from 21 language families in a unified IPA format.
This database is composed of concept aligned word lists for these languages, containing 1016 concepts, each of them translated in most languages. 
However, most of these languages are from Eurasia, hence the collection lacks the typological diversity we would ideally like.
\textbf{Wikipedia}
allows us to investigate a broader and more diverse set of languages, but has no phonetic information (only graphemes) and lexicons extracted from it may be ``contaminated'' with foreign words. We fetch the Wikipedia for a set of 41 diverse languages,\footnote{These languages were: af, ak, ar, bg, bn, chr, de, el, en, es, et, eu, fa, fi, ga, gn, haw, he, hi, hu, id, is, it, kn, lt, mr, no, nv, pl, pt, ru, sn, sw, ta, te, th, tl, tr, tt, ur, zu.} and tokenise their text using language-specific tokenisers from spaCy \citep{spacy2}. When a language-specific tokeniser was not available, we used a multilingual one. We then filtered all non-word tokens---by removing the ones with any symbol not in the language's scripts---and kept only the $10{,}000$ most frequent types in each language.

\begin{table*}[t]
    \centering
    \begin{tabular}{l c c c c c c}
    \toprule
    & & \multicolumn{4}{c}{Surprisal} \\\cmidrule{3-7}
Dataset & \# Languages & Forward & Backward & Unigram & Position-Specific & Cloze \\
    \midrule
CELEX & \phantom{0}\phantom{0}3 & {\color{blue} 3} $\mid$ 0 & {\color{blue} 0} $\mid$ {\color{red} 3} &  {\color{blue} 2} $\mid$ 0 & {\color{blue} 2} $\mid$ {\color{red} 1} & {\color{blue} 2} $\mid$ {\color{red} 1} \\
NorthEuraLex & 107 & {\color{blue} 106} $\mid$ 0\phantom{0}\phantom{0} & {\color{blue} 11} $\mid$ {\color{red} 31} & {\color{blue} 71} $\mid$ {\color{red} 1\phantom{0}} & {\color{blue} 24} $\mid$ {\color{red} 4\phantom{0}} & {\color{blue} 45} $\mid$ {\color{red} 1\phantom{0}} \\
Wikipedia & \phantom{0}41 & {\color{blue} 41} $\mid$ 0\phantom{0} & {\color{blue} \phantom{0}0} $\mid$ {\color{red} 39}  &  {\color{blue} 39} $\mid$ {\color{red} 1\phantom{0}} & {\color{blue} 31} $\mid$ {\color{red} 1\phantom{0}} & {\color{blue} 35} $\mid$ {\color{red} 2\phantom{0}}  \\
    \bottomrule 
    \end{tabular}
    \caption{Number of languages in the analysed datasets with significantly larger surprisals in {\color{blue} initial} $\mid$ {\color{red} final} positions.
    }
    \label{tab:results-significance}
\end{table*}

\section{Experiments and Results}
\paragraph{Forward Surprisal.}

\begin{figure}
    \centering
    \begin{subfigure}[b]{.5\linewidth}
        \centering
        \includegraphics[width=\columnwidth]{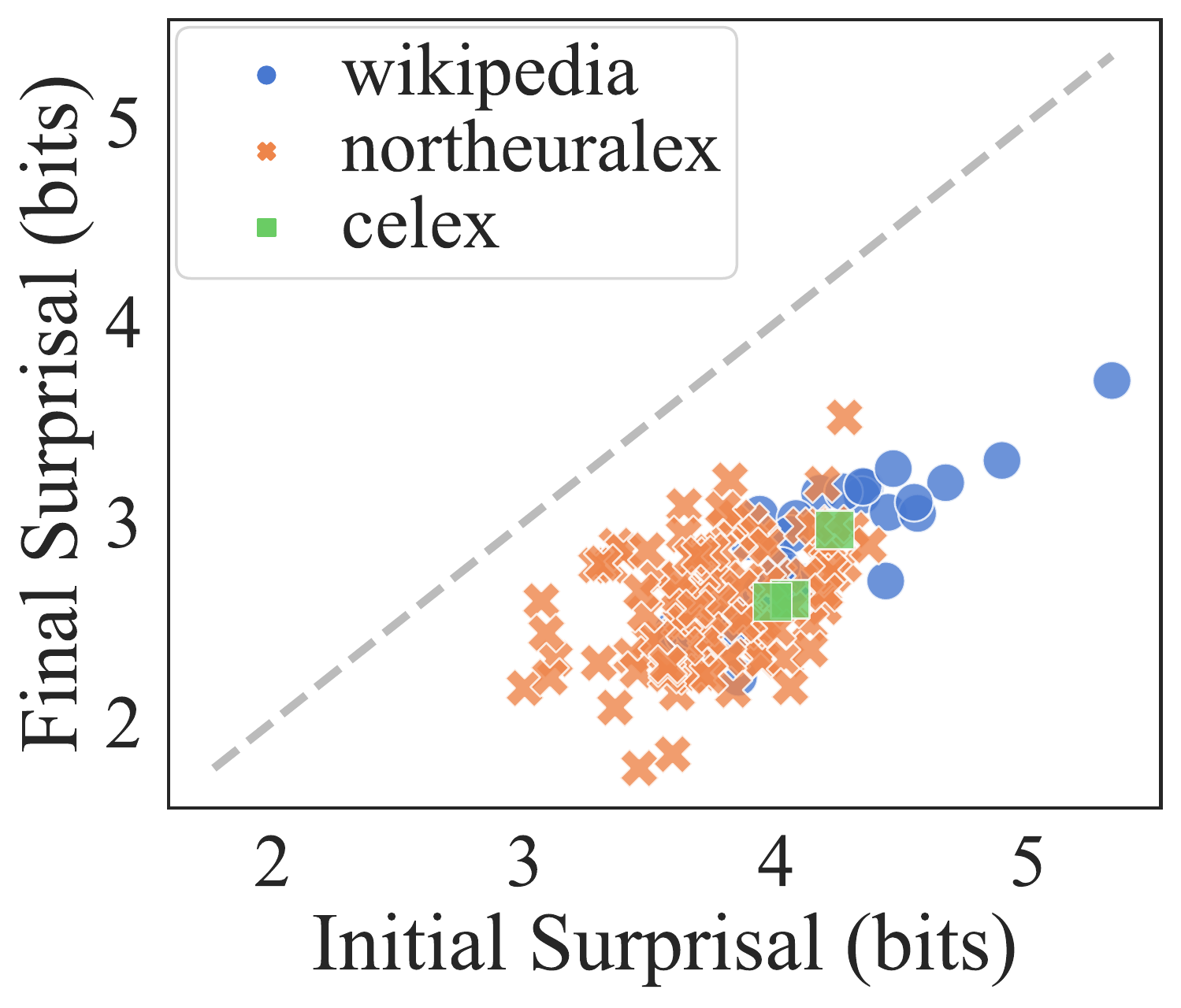}
    \end{subfigure}%
    ~
    \begin{subfigure}[b]{.5\linewidth}
        \centering
        \includegraphics[width=\columnwidth]{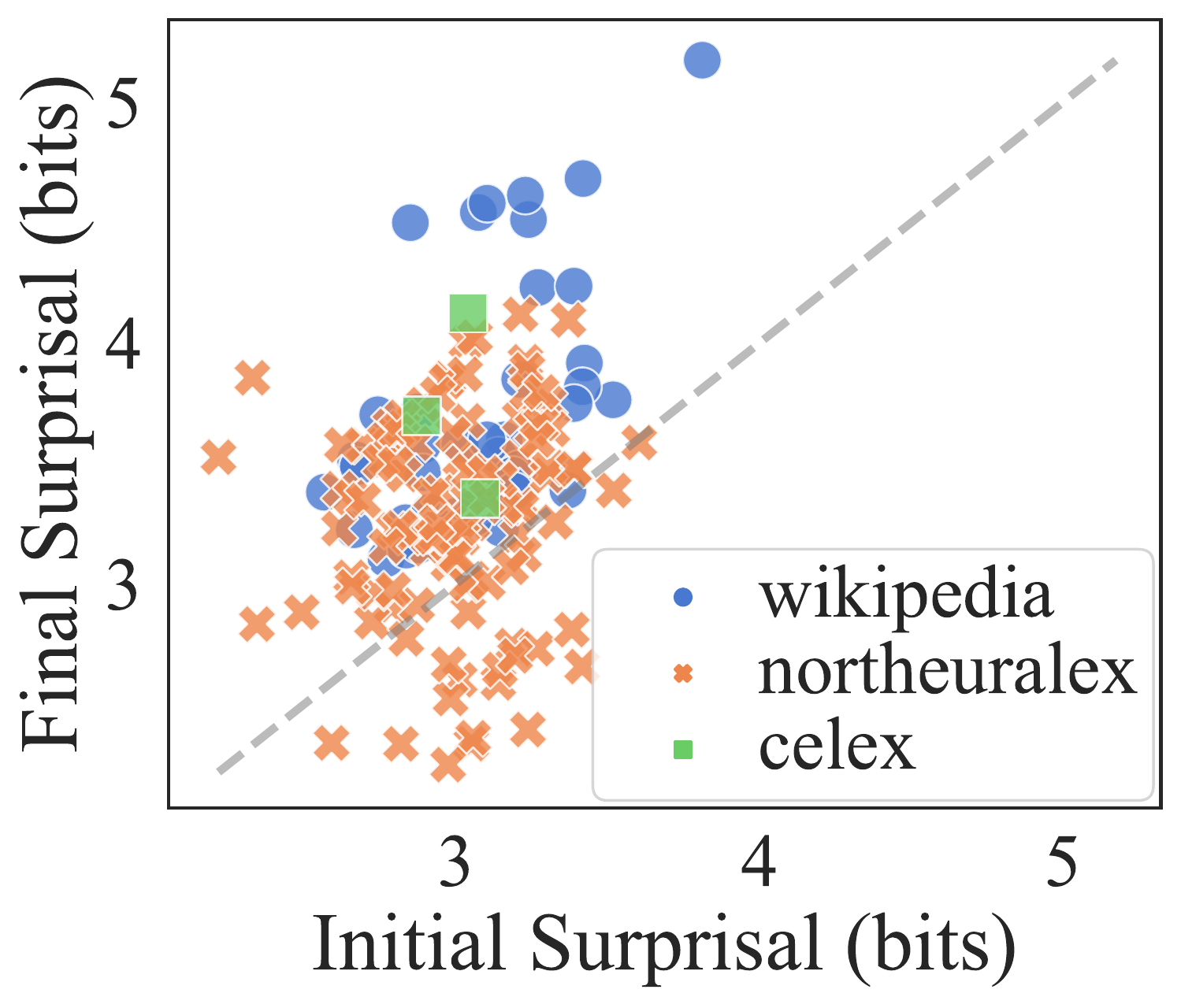}
    \end{subfigure}
    \caption{Word initial vs. final surprisals with: (left) Forward; (right) Backward.}
    \label{fig:binned_lstm}
\end{figure}

We first replicate the results from \citet{son2003information,son2003efficient}, \citet{wedel2019incremental}, and \citet{pimentel2020phonotactic}, which show that surprisal decreases with as the words position advances. 
On average, forward surprisal, i.e. $\enttheta(W_t \mid W_{<t})$, could decrease for two reasons: 
(i) words indeed front-load disambiguatory signals; or (ii) the trivial fact that conditioning reduces entropy.
For each word, we first get the forward surprisal for each segment in it. We then group surprisal values in two groups: word initial (when they are in the first half of its word) and final (when in the second half), ignoring mid positions in words with uneven lengths; 
we average these initial and final surprisals per word, getting a single value of each per word. 
This way we compare earlier vs. later word positions while ignoring any length effect---words with all lengths will possess segments in both groups.
For each analysed language, we then use %
permutation tests (permuting word initial and final surprisals) to evaluate if one group is statistically larger than the other---using $100{,}000$ permutations.
All but one language in three analysed datasets had significantly larger surprisal in word initial positions\footnote{All statistical significance results in this work have been corrected for multiple tests with \citet{benjamini1995controlling} corrections and use a confidence value of $p<0.01$.}---the exception being Abkhaz in NorthEuraLex. These results can be seen in \cref{tab:results-significance} and in \cref{fig:binned_lstm} (left).

\paragraph{Backward Surprisal.}

If the result for forward surprisal is largely due to the amount of conditional information, then reversing the strings should lead to a roughly opposite effect.
With this in mind, for each language, we again bin surprisals in word initial vs. final position, but now we evaluate languages using backward surprisal, i.e., $\enttheta(W_t \mid W_{>t})$.\footnote{We note that \citet{king2020greater} also used backward surprisal, although with a different objective in mind. In one of their experiments, they presented aggregate results of a comparison between the forward and backward surprisal.}
When using backward surprisal, many of the analysed languages have significantly higher surprisals in word final positions (see \cref{tab:results-significance} and the right graph in \cref{fig:binned_lstm}). However, 11 languages in the NorthEuraLex dataset still have higher word initial surprisals, suggesting that initial positions in these languages are indeed largely more informative than final ones.\footnote{We also ran the same experiments with a probabilistic trie model like the ones used in \citet{son2003efficient} and \citet{wedel2019incremental}, which showed an even stronger result reversal when using backward surprisal.}
There does seem to be a large effect of the amount of conditional information and also some lexical effect of front-loading disambiguatory signals, however it is difficult to determine if there are cross-linguistic tendencies with these measures.

\paragraph{Unigram Surprisal.}

To control for the conditioning aspect of the question: \emph{do words front-load their disambiguatory signals?}, we can look at unigram surprisal $\enttheta(W_t)$. This value tells us how uncommon the segments that appear in a certain position are, when analysed in isolation from the rest of the word---uncommon segments are more informative and provide stronger signal for disambiguation.
In NorthEuraLex, 71 of the languages have significantly higher informativity in word beginnings than in endings---nonetheless, one language (Kildin Saami) has higher surprisals in word endings. In CELEX, Dutch and German have higher surprisals in initial positions, but English does not. And in Wikipedia, all languages but Hebrew and Bengali have higher surprisal in initial positions---with Bengali having higher surprisal in word endings.
This experiment suggests that indeed most languages are biased towards providing stronger disambiguatory signals in word beginnings, even when we control for the amount of conditional information. 
Nonetheless, this is not a universal characteristic which all languages share and two analysed languages even had a statistically significant inverse effect.\looseness=-1
\paragraph{Position-Specific Surprisal.}

While cloze surprisal makes explicit the non-redundant informativity a segment conveys, unigram surprisal analyses the same segments in isolation. Position-specific surprisal provides a midway analysis, incorporating the position as some previously-specified knowledge, but not conditioning on the other segments in the word.
The position-specific surprisal is inspired by \citet{nooteboom1988search} experiments, which prime individuals on word length and position. 
As can be seen in \cref{tab:results-significance}, position-specific surprisal again seems to favour initial positions over final, but only slightly. 
Interestingly, most languages present no significant difference and some the inverse effect (i.e. higher surprisal in final positions).\looseness=-1
\paragraph{Position-specific Unigram models.}
To better understand the differences between the unigram and position-specific surprisal results, we trained position-specific unigram models---which count each segment's frequency per position---and then calculated their Kullback--Leibler (KL) divergence per position with the traditional unigram%
\begin{align}
    \KL(p(w_t \mid t) &\mid\mid p(w_t)) \\
    &= \sum_{w_t \in \Sigma} p(w_t \mid t) \log \frac{p(w_t \mid t)}{p(w_t)} \nonumber
\end{align}
We compare these KL divergences and find that, for all but four  languages, the KL is largest in either the first or second segment positions.%
\footnote{We use Laplacian smoothing in the position-specific unigrams and constrain the analysis to positions which appear in at least 75\% of the analysed words in that language.} 
This suggests that one of the reasons for higher unigram surprisal in initial positions is that the first two segments usually differ from the rest of the positions, potentially serving as markers for word segmentation.

\paragraph{Cloze Surprisal.}

\begin{figure}[t]
    \centering
    \includegraphics[width=.85\columnwidth]{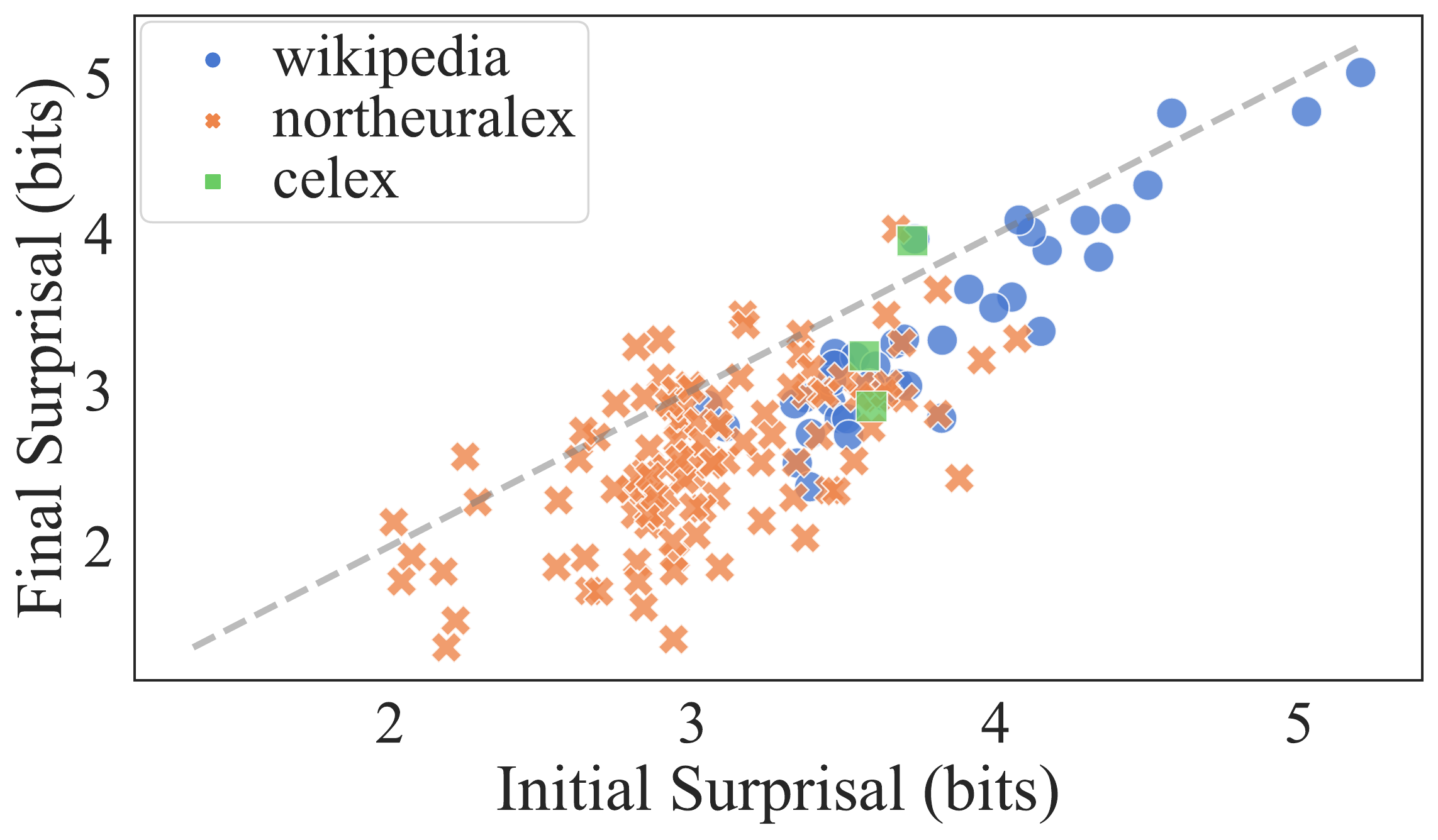}
    \caption{Word initial vs. final cloze surprisals.}
    \label{fig:binned_cloze}
\end{figure}

When we condition a segment on all others in the same word, we measure how much uncertainty is left about that individual segment when considering everything else, or, in other words, how much information is passed only by that segment non-redundantly.
Word initial surprisal is higher in most analysed languages (see \cref{tab:results-significance}). 
Nonetheless, two languages in Wikipedia, Thai and Bengali, have significantly higher surprisal in their final segments---while English in CELEX and Hungarian in NorthEuraLex also present this same inverse effect. 
Front-loading disambiguatory information, thus, is not established to be the linguistic universal it is believed to be, with only roughly half the analysed languages showing this property when we control for morphology (CELEX and NorthEuraLex).
\cref{fig:binned_cloze} plots the results for all languages analysed.
When we compare these results, we find an interesting pattern.
Morphology seems to reduce non-redundant (cloze) information later in the words---while only half of the languages had significant surprisals in CELEX (which consists of monomorphemic words) and NorthEuraLex (base forms), most languages were significant in Wikipedia. Furthermore, English and Hungarian had significantly higher surprisals in word endings in CELEX and NorthEuraLex, while the opposite trend in Wikipedia---this is consistent with the fact that suffix morphemes are present in more types than word roots are, so morphology would make word endings less surprising.
\paragraph{Length as a Confounding Effect.} \label{sec:study_length}

We evaluate the impact of length as a confounding effect on previous methodologies.
As mentioned in \cref{sec:critique}, by directly analysing surprisal--position pairs (as opposed to binning word initial vs. final positions), previous work confounds position and word length---i.e., only long words will have later word positions.
In this study, we analyse forward surprisal--length pairs; instead of pairing a segment's surprisal with its position, we pair it with its word length. We then get the slope formed by a linear regression between these pairs of values and test for its significance per language by using a permutation test, in which we shuffle surprisal--length values.
On the three datasets, all languages have statistically significant negative slopes, meaning long words have smaller surprisals on average than shorter ones.%
\footnote{\newcite{king2020greater} indeed present a similar correlation in their Figure 2.}
A caveat, though, is that now we are confounding position into our length analysis.
Constraining our analysis only to the first two segments in each word, we still find the same effect---though now one language (Hebrew) in Wikipedia and seven in NorthEuraLex are not significant.
We can thus conclude that longer words have smaller surprisal values than shorter ones, even when controlling for the same word positions. 
This implies that directly using surprisal--position pairs for such an analysis is not ideal.

\begin{table}
    \centering
    \small
    \begin{tabular}{l c c c c c c}
    \toprule
    & \eow{} & Non-\eow{} \\
    \midrule
Forward & 1.14 & 3.55 \\
Backward & 0.89 & 3.61 \\
Unigram & 2.75 & 4.90 \\
Position-specific & 0.00 & 4.36 \\
Cloze & 0.00 & 3.23 \\
    \bottomrule 
    \end{tabular}
    \caption{\small Average surprisal (in bits) of \eow{} vs. non-\eow{} segments averaged over all datasets. 
    }
    \label{tab:results-eow}
\end{table}

\paragraph{The Effect of End of Word in Surprisal.} \label{sec:eow}
The end-of-word (\eow{}) symbol is a special ``segment'' which symbolises the end of a string. It is necessary when modelling the probability distribution over strings $\bw \in \Sigma^*$, to guarantee that the overall distribution sums to 1. Nonetheless, it is expected to behave in a different way from other segments.
If a speaker wants to reduce their production effort, although changing from one phone to another may help, the most efficient way is usually just ending the string earlier. Furthermore, since all realisable strings must eventually end, it will be present in all words, making it a very frequent symbol---in fact, \cref{tab:results-eow} shows its average surprisal is much lower than that of other segments. As such, it is only natural it should be analysed on its own, separately from other segments. Through the same logic, other segments should also be analysed separately from \eow{}---or else, lower word final surprisals may be due to this symbol alone.
As such, we analyse the surprisal of LSTM ``language models'' without the \eow{} symbol here.\footnote{To be more precise, we actually ignore the beginning-of-word symbol when estimating backward surprisal.}

Unsurprisingly, \cref{tab:results-diffs} shows the difference between word initial and final positions is considerably reduced when we remove the \eow{} symbol from the forward surprisal analysis. Surprisingly, we see that when we remove the beginning-of-word from the backward surprisal analysis, instead of a larger word final surprisal, we get a larger word initial value---even though we are still conditioning the models right-to-left. This result further supports the hypothesis that the disambiguatory signals are on average stronger in word initial positions.

\begin{table}
    \centering
    \resizebox{\columnwidth}{!}{%
    \begin{tabular}{l c c c c c c}
    \toprule
     & \multicolumn{3}{c}{\eow{}} & \multicolumn{3}{c}{No \eow{}} \\\cmidrule(lr){2-4} \cmidrule(lr){5-7}
    & Initial & Final & Diff (\%) & Initial & Final & Diff (\%) \\
    \midrule
Forward & 3.85 & 2.65 & 31.1 \% & 3.83 & 3.00 & 21.6 \% \\
Backward & 3.02 & 3.40 & -11.3 \% & 3.63 & 3.39 & 6.7 \% \\
Unigram & - & - & - & 4.85 & 4.40 & 9.3 \% \\
Position & - & - & - & 4.36 & 4.17 & 4.3 \% \\
Cloze & - & - & - & 3.26 & 2.81 & 13.9 \% \\
    \bottomrule 
    \end{tabular}
    }
    \caption{\small Average surprisal per segment in word initial and final positions with and without \eow{} symbols.
    }
    \label{tab:results-diffs}
\end{table}

\section{Conclusions}

In this work, we analysed the distribution of disambiguatory information in word positions. 
We present an in-depth critique of previous work, showing several confounding effects in their analysis.
We then proposed the use of three new methods which corrected for these biases---namely unigram, position-specific and cloze surprisal. 
These models controlled for the amount of conditional information across word positions, allowing for an unbiased analysis of the lexicon.
Using these models we show that the lexicons of most languages indeed front-load their disambiguatory signals. 
This effect, though, is not universal and the difference in disambiguatory information between word initial and final positions is much lower than previously estimated---ranging from 4\% to 14\%, depending on the used metric, instead of 31\%.

\bibliography{tacl2018}
\bibliographystyle{acl_natbib}

\end{document}